\documentclass[10pt, twocolumn, letterpaper]{article}
\usepackage{iccv}
\usepackage{times}
\usepackage{epsfig}
\usepackage{graphicx}
\usepackage{amsmath}
\usepackage{amssymb}
\usepackage{bm}

\usepackage{subfigure}
\usepackage{multirow}
\usepackage{array}
\usepackage{authblk}

\usepackage[pagebackref=true,breaklinks=true,letterpaper=true,colorlinks,bookmarks=false]{hyperref}

\iccvfinalcopy 

\ificcvfinal\fi

\makeatletter

\newcommand{\Rmnum}[1]{\expandafter\@slowromancap\romannumeral #1@}
\makeatother

\begin{document}

\title{UU-Nets Connecting Discriminator and Generator for Image to Image Translation}

\author{Wu Jionghao\\
National University of Singapore\\
21 Lower Kent Ridge Rd, Singapore 119077\\
{\tt\small E0046486@u.nus.edu}}

\maketitle

\begin{abstract}
Adversarial generative model have successfully manifest itself in image
    synthesis. However, the performance deteriorate and unstable,
    because discriminator is far stable than generator, 
    and it is hard to control the game between the two modules. 
    Various methods have been introduced to tackle the problem such as WGAN \cite{gulrajani2017improved:WGAN}, 
    Relativistic GAN \cite{jolicoeur2018relativistic:RSGAN} and their successors by adding or restricting the loss function, 
    which certainly help balance the min-max game, 
    but they all focused on the loss function ignoring the intrinsic structure limitation. 
    We present a UU-Net architecture inspired by U-net bridging the encoder and the decoder,
    UU-Net composed by two U-Net liked modules respectively served as generator and discriminator. 
    Because the modules in U-net are symmetrical, therefore it shares weights easily between all four components.    
    Thanks to UU-net's modules identical and symmetric property, we could not only carried the features from inner generator's encoder to its decoder, 
    but also to the discriminator's encoder and decoder. 
    By this design, it give us more control and condition flexibility to intervene the process between the generator and the discriminator.

\end{abstract}


\vspace{-8pt}
\section{Introduction}
\vspace{-5pt}

Latterly there has been considerable attention on generative models for the task of image or video information
    synthesis, segmentation and translation \citation {isola2017image:pix2pix2016, peopleInClothing, 
    PoseGuidedGeneration, igan, cycleGAN}.  
    The state-of-the-art models achieved success in considerable computer vision tasks.
    Many of them are variants of the basic 
    encoder-decoder neural networks like Autoencoder \cite{kingma2013auto:VAE} and U-Net \cite{ronneberger2015u:UNet}.
    \begin{figure}
    	\begin{center}
            \includegraphics[width=0.45\textwidth]{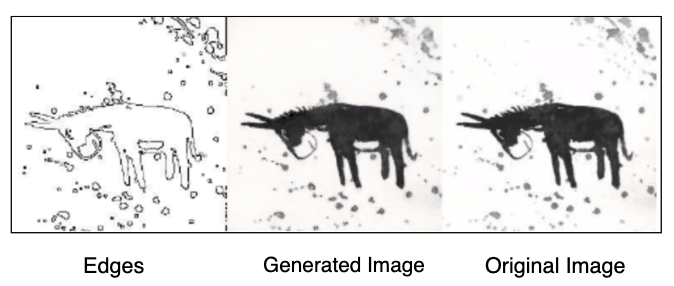}
    	\end{center}
      \caption{\small{Generated Chinese painting from the master JIBAISHI
      .}}
    	\label{fig:teaser}
    \end{figure}
    These encoder-decoder networks share a fundamental similarity:
    skip-connections, which carry the high-level and meaningful features from the decoding sub-network with low-level and superficial features from the encoding sub-network together for the final output. Till now, skip-connections have proved solid in
    restoring fine-grained information of the target images. As an illustration,  instance-level segmentation, namely U-Net,
    which empower the segmenting of blocked objects. Later combining with Generative Adversarial training 
    process, like VAE-GANs\cite{larsen2015autoencoding:vaegan}, CVAE-GANs \cite{bao2017cvae:cvaegan} could utilize the discriminator to enhance the autoencoder's performance through the minimax game.
    Doubtlessly, image translation in general has reached a astonish level of performance.
    And these U-shape GAN architectures tend to be mature, 
    through these years many components have been added to the backbone network of the GAN architecture, 
    like bigan \cite{donahue2016adversarial:BiGAN}, Cycle-GAN\cite{zhu2017unpaired:cycleGAN} changed the structure via creating branches; 
    ACGAN \cite{odena2017conditional:ACGAN}, Bycicle GAN \cite{zhu2017toward:bicyclegan} adds additional components which inserting one or more classifiers or other functional Nets; 
    Info GAN, Glow and VAE-GANs restrict or manipulate the hidden space by mathematics; 
    Multi-GAN, Stack-GAN \cite{zhang2017stackgan:Stackgan} reuse or use more generators or discriminators.

    But does the main structure reached limitation? Or there is no space to change the main structure of generative adversarial network?

    Researchers reveal that there are some common identified problems in GAN structures. 
    Mode collapse and unfair training process between the two contradictory players are the two major critics. 
    In this paper, we mainly investigate the potentiality of our UUNets to deal with the second problem of instability and uncontrollability between generator and discriminator.

    In this paper, we suggested two changes, 
    one is to design both generator and discriminator as U-Net architecture. 
    Another is using skip connection to bridge generator and discriminator and back-propagate dependently.

    First, as we know that recent years there are many changes in U-shape architectures, 
    Some may change the naive U-Net to densely connected by changing the connection topology within the U-shape networks such as UNet++ \cite{zhou2018unet++:Unet++}.
    CU-Net \cite{tang2018cu:CUNet}, and Stack-Unet \cite{shah2018stacked:stackunet} stack one or more U-Net together. After that they could use skip connection in layers of encoders and decoders differently. 
    Through the development of UNet we have an very important observation which is the intrinsic similarity or symmetric property within Unet's encoder and decoder.
    This nicely formed architecture allow components easily concatenates with the similar parts. 
    It is inspired us with the UUNet design, 
    which embraced the idea of symmetric design consisted with two Unets one for generator and one for discriminator. 
    In other words, two encoders and two decoders they all similar and could form many symmetrical patterns that set the foundation for various UUNet-GAN.
    Then UUNet extends U-Net skip connecting the inner part of generator and discriminator in many ways formed the first family of UUNets.
    
    Second, in order to see the advantages of UUNet design we have to go back to the second common problem, 
    the unbalanced game of the two components. 
    First, instability comes from the original design. 
    Two individual function body perform the minimax game, 
    and Generator tries to deceive Discriminator, one the other hand, the discriminator ties to identify the fake out produced from the generator. 
    During the training process, Generator conduct an image translation problem and the Discriminator conduct a binary classification problem, 
    then based on their performance backpropagate independently. 
    In many observations, discriminator is much stronger than the generator and reach the convergence much quicker. 
    As a result, generator is hard to learning the information from the game. 
    We propose the UUNet GANs adding skip connection between Generator and discriminator. 
    Therefore during the train, they backpropagate the gradient to each other or simply from one to the other. 
    In this architecture, generator could absorb gradient from discriminator, 
    in other words obtain information while training discriminator. 
    What’s more, we could design many more connection and adding modules to intervene the process. 
    In conclusion, we could train the two previously independent part dependently.

To sum up, we have following contributions:
\begin{itemize}
    \item To the best of our knowledge, we are the first connecting the inner components of the discriminator and generator for GAN models;
    \item Introduced a symmetric UU-Net framework that allow different ways to connect components, to control and to balance the unfair minmax game between generator and discriminator;
    \item Adds loss function between discriminator and generator via UUNet-VAE-GAN framework.
    \item UUnet-GAN can leverage the similarity structure between generator and discriminator, and use additional restriction between the two components.
    

\end{itemize}


\section{Related Work}
In this section, the content falls into three categories. 
The overall image translation tasks, current algorithms and generative models and our proposed UUNet framework.

\subsection{Generative Adversarial Networks (GANs)}

As one of the most promising generative models, Generative Adversarial Network (GAN) \cite{goodfellow2014generative:GAN} is 
studied extensively recently. The mission of GAN is to approximate a
generator distribution $P_g(x)$ that matches the real data distribution $P_{real}(x)$ through a mini-max game:

\begin{equation}
\begin{split}
	\mathcal{L}_D =	\min_G\max_D & \{ \mathbb{E}_{x \sim {p_{real}}}[\log D(x)]  \\
  &+  \mathbb{E}_{z \sim { p_{z}}}[\log (1 - D(G(z)))]\}	
\end{split}
\end{equation}

\begin{equation}
\mathcal{L}_G =	-\max_D  
    \mathbb{E}_{z \sim { p_{z}}}[\log D(G(z))]
\end{equation}

Basic GAN is not only unstable and uncontrollable, 
but also hard to choose and explain the prior noise z. 
To tackle these shortcomings, a various of GAN have
emerged, such as conditional GAN (cGAN) \cite{mirza2014conditional:CGAN}, VAE-GAN, relativistic
GAN (RS-GAN) \cite{jolicoeur2018relativistic:RSGAN} and T-GAN (Turing GAN) \cite{su2018training:TuringGAN}. cGAN tries to solve
the controllability issue via providing class labels to both generator
and discriminator as condition. VAE-GAN aims to regularize the hidden space
more interpretable representation via maximizing the mutual
information between hidden code and generator output and also change the 
GAN pattern from map randomly selected prior space z to x' into meaningful input original image x to generated image, which also 
adds additional function to control input with the output 
by adopting this methods a lot of new use case has been created through the past years, 
for example raindrop removing, dehaze and impainting tasks. Last but not least, another 
big issue with the unstable training process are mitigate by various methods. WGAN introduces earth mover distance to measure similarity
of real and fake distribution, which makes training stage more
stable. Our work is related to VAE-GAN and UNet-GAN closely. And details are discussed in the sequel.

\subsection{U-like Algorithms}

The beginning U-Net architecture was invented by Ronneberger et al, and outperform other algorithms in segmentation of cells in bio-medical images using small amount of labeled data, while pix2pix was the one used U-net achieved state of art performance with the task of image to image translation. The most important operations in U-like structure are up-sampling and skip connection concatenation of the modules between encoder and decoder respectively.

First, Long et al designed fully conventional networks (FCN) \cite{long2015fully:FCN}, up-sampled feature maps are produced with transpose convolution operation, which automatically generate the symmetrical inverse mapping for the convolution. This makes it easy to restore small hidden information to the original input tensor shape. Second, feature maps skipped from the encoding part, then join the decoding part of U-Net and add convolutions with non-linearity function between each up-sampling step. Therefore, 
the topology of the U-net structure takes the context information at multiple scales and propagate the abstract layers directly to the concrete layers.  
 The skip connections was used in many place like res-net and has been adopted in many fashion, mainly solving gradient vanish for very deep architectures and enhancing the reconstruction from rather small hidden space via it maintain the information in shallow layers to the deeper layers. 2018 a variational U-Net (U-VAE) illustrate the idea with decent performance on appearance and shape generation by nicely combining variational auto-encoder with U-Net design.
 
 The skipped connections have not only demonstrated the power of recovering the spatial resolution and it also been a great architecture topology design concept. For example, DenseNet \cite{huang2017densely} utilize the skip connection pattern making every module connecting together benefiting in sharing information via building relationships between layers. U-Net++ has intensive experiments with different topology, finally it embraced a Pascal's triangle architecture. it is showing that a nicely constructed structure with mathematical beauty could reveal the nature of regularity.

\subsection{Image-to-Image Translation}

Image-to-image translation is a general problem which goes back at least to Hertzmann et al.’s
Image Analogies \cite{hertzmann2001image:imageAna}. Lots of problems can be thought as sub-problem of this kind, such as segmentation, impainting, etc.
Recently pix2pix which achieved great result on the general problem. Inspired by this work,
we builds on our UUNet framework, 
which also double the U-shape network, one for learning a mapping from input to
output images one for the codification. 

\subsection{Traditional Chinese Painting Imitation}
 Chinese painting translation between the learner's painting to the master's is  also applied with our UUNet GAN model. We use model's to generate the imitation of the master Jipaishi painting and the other way around. The task is asking the model to learning the mapping between a naive painter's work and the master's work, so that we use the mapping to guide young painter to create new art.

\begin{figure}
    \begin{center}
        \includegraphics[width=0.45\textwidth]{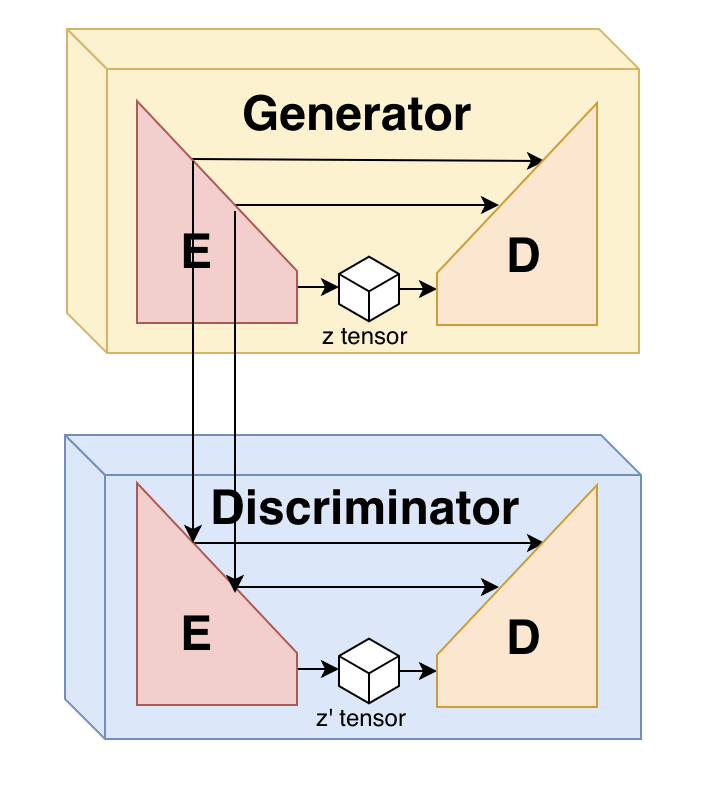}
    \end{center}
 \caption{\small{Our models formed an UU-shape like family of models respectively serve as Generator and Discriminator, the main structure is original GAN design.
 %
    \label{fig:teaser}}}
\end{figure}
\section{Methodology}
In this section we present our UUNet-GAN framework. First and foremost, we will give an overview of our problem and architecture, and 
then illustrate different possibility of UUNets connections with figure3. Finally, 
Based on various connections we will demonstrate how UUNets GAN family could allow communication between discriminator and generator during the process of training.

\begin{figure*}
	\begin{center}
    		\includegraphics[width=0.95\textwidth]{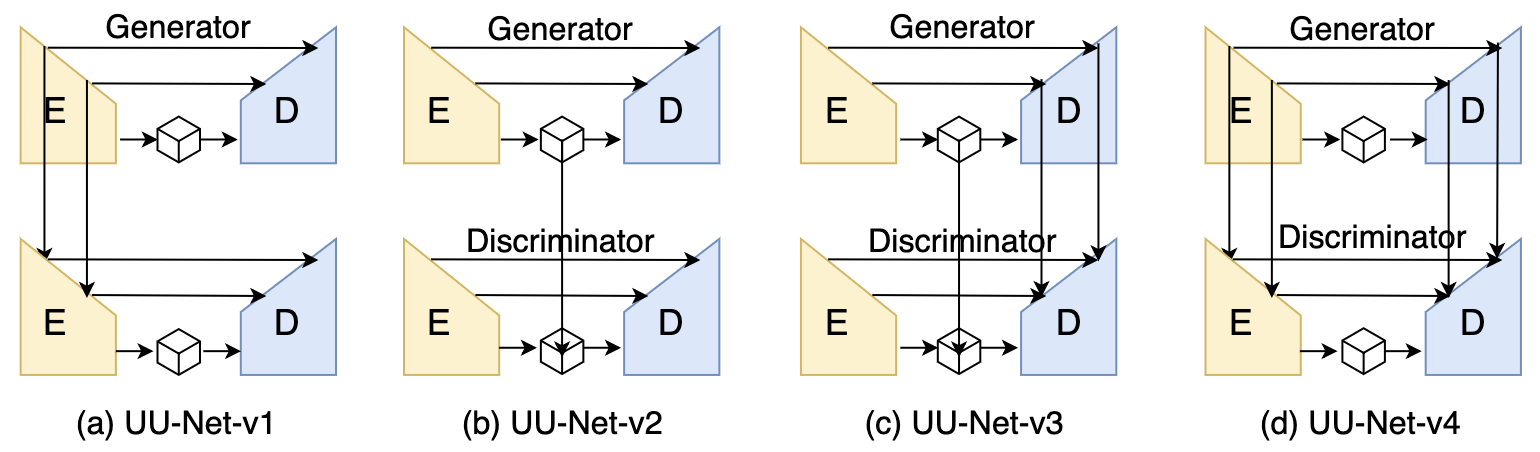}
    	\end{center}
      \caption{\small{Our UU-Net family was constructed under the same protocol: two U-shape like modules respectively serve as Generator and Discriminator, the difference is the way how they connect to each inner components.
          	\label{fig:teaser}}}

\end{figure*}
\subsection{Overview}
As discussed in previous section there is an unfair game between the generator and discriminator,
We put a way to solve it by adding connections between the opponents. In order to connect them, we
draw inspiration from U-Net, we further form our network as two U-shape network since it share the nice property, symmetric.

\subsection{Network Architecture}
Fig. 2 shows a high-level overview of one example of the suggested UU-Net architecture. 
As seen, UU-Net composed with two similar subnet as an encoder sub-network or backbone followed by a decoder sub-network. 
What distinguishes UUNet from other U-Net or GAN design is two fold. the connection between generator and discriminator and 
the nicely designed double U-shape, which is easy and nature for various skip connection and also compatible with other developments of U-Net.

We also tried combining our UUNet framework with VAE network in order to fully utilize and demonstrate our connection mechanism, which could form more condition with two variational hidden spaces, such that achieving our goal of control the unbalance and the race between generator and discriminator. 

\subsection{Tail module}
Tail module as show in fig. 4 is designed, because we find out through later experiments that it is too sudden if asking the discriminator U-net's decoder output directly down to one dimension. So we add an additional tail module for smooth transition from input image dimension to final binary output dimension. In later discussion we find out that the tail module could also serve as a shock absorber which help calm down the unstable training process and increase the overall performance.

\subsection{Formulation}
Prior studies have found it constructive to merge the GAN objective with reconstruction loss in VAEGAN. 
We inherent with the traditional GAN loss.
The discriminator is perform like a functional object which enhance the reconstructing functionality of the auto-encoder, 
On the other hand, the generator is targeted not only to deceive the discriminator but
also to be closely located at the neighborhood of ground truth manifold in an L2 measurement.

But to utilize our UUNets architecture and 
implement the idea of intervening the two contradictory body and also mitigating the discriminator 
through letting the generator get the information from discriminator during each training step, 
we introduce another possibility to form penalty functions.

Following the initial VAEs~\cite{kingma2013auto:VAE}, 
we pick the centered isotropic multivariate Gaussian \(N(0,I)\) as the prior \(p(z)\) over the hidden variables. 
The encoding network \(E\) is aim to produce 
two independent vector, \(\mu\) and \(\sigma\).  Therefore the posterior could be formulate as \(q_\phi(z|x) = N(z; \mu, \sigma^2)\). 
Thus the Decoding network \(D\) receives the hidden vector z, which is sampled from \(N(z; \mu, \sigma^2)\) using the technique of reparameterization: \(z = \mu + \sigma \odot \epsilon\) where \(\epsilon \sim N(0,I)\). 

With the KL-divergence as the distance between distributions, regression loss, \(L_{G_{kl}}\) (i.e., \(E(x)\), given \(N\) data samples, could be quantified as follows:
\begin{equation}\label{e9}
  \mathcal{L}_{G_{kl}}(z;\mu, \sigma) =  \frac{1}{2} \sum_{i=1}^{N} \sum_{j=1}^{M_z} (1+ \log (\sigma_{ij}^2) - \mu_{ij}^2 - \sigma_{ij}^2)
\end{equation}
where \(M_z\) is the dimension of the hidden vector \(z\).

As to the reconstruction loss, \(L_{RE}\), we choose the widespread use pixel-wise mean squared error (MSE) function. Let \(x_r\) be the reconstruction sample, \(L_{RE}\) is defined as:
\begin{equation}\label{e10}
  \mathcal{L}_{RE}(x, x_r) =  \frac{1}{2} \sum_{i=1}^{N} \sum_{j=1}^{M_x} \| x_{r,ij} - x_{ij} \|^{2}_{F}
\end{equation}
where \(M_x\) is the dimension of the data \(x\).

\subsubsection{Loss Function on Generator's and Discriminator's Hidden Space}
To restrict the hidden output from discriminators' encoder closely 
to similar position of the hidden output of generator's encoder's manifold. we used the cross KL-divergence \(L_{CKL}\), which means we optimizes the distance between generator's posterior \(q_{G\phi}(z|x) = N(z_{G}; \mu_{G}, \sigma_{G}^2)\) and the
discriminator's posterior distribution \(q_{D\phi}(z|x) = N(z_{D}; \mu_{D}, \sigma_{D}^2)\).
For distance can be computed as below:
\begin{align}
\begin{split}
\mathcal{L}_{CKL}(G,D)
&= \log \frac{\sigma_G}{\sigma_D} + \frac{\sigma_D^2 + (\mu_D - \mu_G)^2}{2 \sigma_G^2} - \frac{1}{2}
\end{split}
\end{align}


Since discriminator could judge real and fake data, here we could set another group of hyper-parameter for controlling the proportion between the real and the fake as to the discriminator's KL loss contribute:

\begin{align}
\begin{split}
\mathcal{L}_{D_{kl}} = \frac{\alpha \mathcal{L}_{KL_{real}} + \beta \mathcal{L}_{KL_{fake}}}{\alpha + \beta}  
\end{split}
\end{align}

where \(\alpha\) and \(\beta\) are the hyper-parameters of controlling the relative importance of real and fake data as to discriminator VAE network.


\subsubsection{Total Loss Function}
To form our final objective function, we just sum the two function with option condition draw from different UUNets and average over batches.

\begin{equation}\label{e9}
  \begin{split}
\mathcal{L}^{(m)} = \arg\min_{\mathbf{G}} \max_{\boldsymbol{D}}  \frac{1}{M}\sum_{i=1}^{M}(\lambda_{dis}\mathcal{L}_{D_{kl}}^{(i)} + \lambda_{G_{kl}}\mathcal{L}_{G_{kl}}^{(i)}
\\ +  \lambda_{re}\mathcal{L}_{RE}^{(i)} 
+  \lambda_{ckl}\mathcal{L}_{CKL}^{(i)} 
)
\end{split}
\end{equation}


where $M$ is the mini-batch size, the superscript $(i)$ denotes the $i$th data point and $\mathcal{L}^{(m)}$ is the loss for the $m$th mini-batch, \(\lambda_{dis}\) and \(\lambda_{re}\) and \(\lambda_{ckl}\) are the hyper-parameters take care of each term's importance.

\subsection{Implementation Details}

\begin{figure*}
	\begin{center}
    		\includegraphics[width=1.0\textwidth]{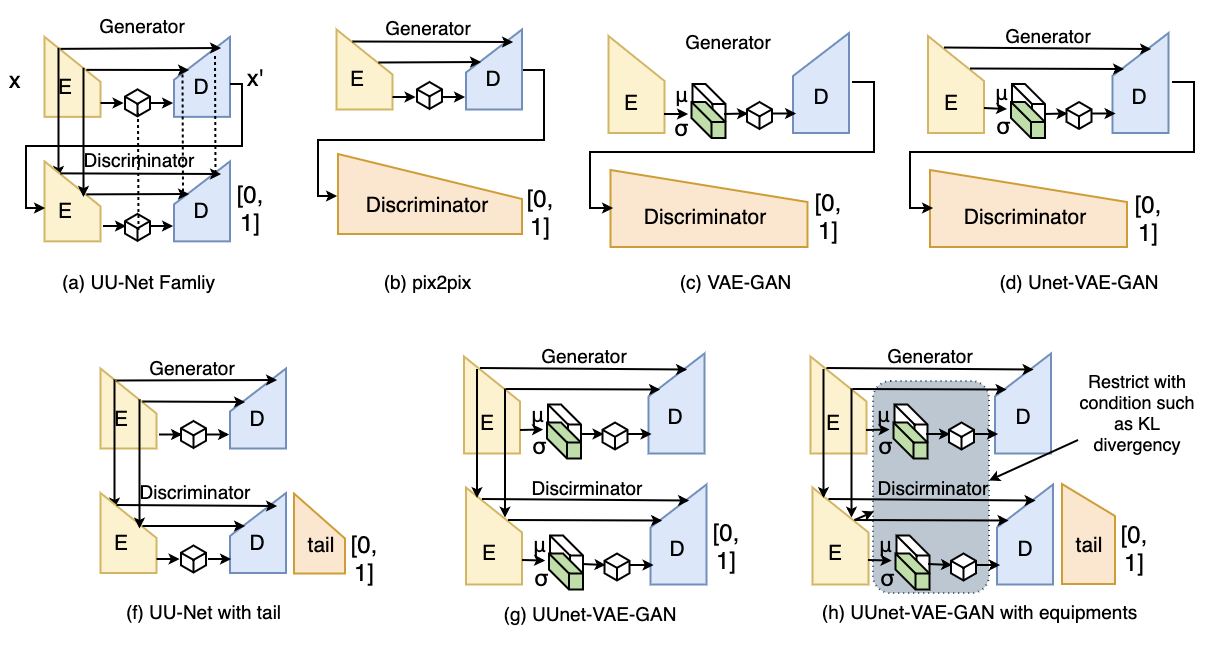}
    	\end{center}
      \caption{\small{Comparsion between different models' architecture including our UUNet-GAN group, UUNet-VAE-GAN group and other classic frameworks.
          	\label{fig:teaser}}}

\end{figure*}

We develop our linked generator and discriminator architectures
from the original Pix2pix design. The following is the discuss of the core ideas and features. More detail can be find in supporting documents.

Our network architecture is demonstrated in Figure 1. It consists of two symmetric U-shape like nets, namely generator and discriminator. For each U-Net, we keep the tradition which shrink the image first (yellow part) and reverse it back to the initial dimension (blue part). The typical architecture of encoder and decoder convolutional networks are composed of multiple replicative
block of 3x3 convolutions, then followed by
a non-linear function (ReLU) and a 2x2 max pooling operator with stride 2 for shrinking. Every shrinking step for generator we make it twice the number of feature channels. At each step, we make up the decoding network of an up-sampling of the feature map followed by a 3x3 de-convolution that inverse the convolution operation. For generator's decoder's concatenation with the correspondingly feature map from the encoder, we combine two 3x3 convolutions, one from encoder and one from decoder. On the other hand, for discriminator's encoder we also concatenate the layer's from generator. As to discriminator's decoder, we have different skip connection strategies, for example, we could use the "triple concatenation" meaning that the discriminator's decoder combine both discriminator's encoder and generator's encoder's feature maps with its own outputs correspondingly under the prerequisite of UUNet's symmetric design. At the final layer of discriminator a convolution with 1x1 kernel is used to shrink the features to the binary space. 

To allow the skip connection between the two independent networks, we use a very standard pattern of 
object-oriented programming. During training we past the entire generator object to the discriminator. 
So the discriminator could use the weights and layers from generator to perform skip connection and during the training it could let the generator "steal" the information. 
While backbagation, the discriminator could backbagate the gradients to the inner instance of generator, such that the 
generator could benefit even when training discriminator. After all, we could balance the race between discriminator and generator.

\subsection{Discussion}
To highlight the novelty of our method, we compare UUNet-GAN framework
with the following Networks. In Fig. 3, we show their
network structures as well as ours for visual comparison.
Pix2Pix learns a generation network G which also serves as an inference network
E. However, in practice, it only uses U-Net for the reconstructions. As we all known, U-Net has some many alternatives, so we also build a VAE-GAN model with skip-connection for the comparison. Our model gradually change the models to our UUNet-GAN design, such as UUnet and UUVAE-GAN to investigate all possibilities of intervene the unbalance training process.

\section{Experiments}
In this section, we formed a family of UU-Net GAN variants. Then we conduct several ways to investigate our new designs, because there are many methods and structures to connect between the two components, namely generator and discriminator.

\subsection{Datasets and Settings}
 
To examine the universality of ours UUNetGANs framework, we assess
the idea on varieties of datasets, including both
semantic segmentation on the Cityscapes
dataset \cite{cordts2016cityscapes} and image translation on our special traditional Chinese Painting dataset, which can be used for 
novice to master painting generation, guiding student coloring 
their paints or just in-painting from old paints.
Semantic segmentation task trained on Architectural labels photo Facades\cite{tylecek2012cmp:facade}. We trained Map aerial photo\cite{isola2017image:pix2pix2016} scraped from Google Maps.
Details of the process of those datasets are provided in supplemental materials. In all instances, 1-3 channel images are provided for experiments. Qualitative results are shown in Figures 5. 
We learn from recent Google's BigGAN \cite{brock2018large:biggan} project that in order to obtain decent results from generative adversarial model, we could set a larger batchsize for the purpose of absorbing more information from the batch all in one step of training. To make a fair comparison, all results
shown in table 1 were trained with 2000 epoch with batch-size of 8 and took two days of training
on four GeForce GTX 1080 Ti GPU in parallel. At test time, all models should also run on four GPU with the same environment setting so as to get the nice results.

\subsection{Evaluation metrics}

How to and what to evaluate the quality of synthesized images remains as an open and interesting question. 

Here we adopted several metrics for comparison in order to get more holistically understanding from different aspect to both the visual quality and the varies of our architecture. Mean square error (MSE) usually ignore in image synthesis problems
because it can not capture the distribution information, but as a tool for understanding the difference as to various part of our intra-UU-Net family modules, we still include it. Peak signal-to-noise ratio (PSRN) is another  
effective quantitative measurements. Finally SSIM is widely accepted in this image structure similarity comparison scenario.
Nevertheless, since the UU-Net structure remain plenty of rooms for more depth research, so here we not merely compare the performance on specific task but also focused on how and what to connect generator and discriminator, so that we could achieve good performance as well as solve and intervene the unbalanced race between the two components. 

\subsection{Baseline models}
For comparison, we used the original U-Net and pix2pix as our baseline. We chose U-Net because it is the root of our design. Then we introduce our a UU-Nets with the same
U-Net generator and a slightly changed U-Net discriminator. This was to make sure that the results of our architecture is not merely owing to adding layers. 

\subsection{Compared Methods}
\subsubsection{UU-Net-GANs Family Comparison} 
UU-Net GAN as we discussed have several variants due its similarity propriety. The similarity has two fold. On one side, the generator and the discriminator are design both in U shape networks. On the other side, inside the main two components, the four encoder and decoder are symmetrical formed. Therefore, the topology is naturally generated, we can either connecting discriminator to generator only on the two encoder part or decoder part separately, or we could connecting both discriminator's encoder and decoder with the generator's encoder and decoder. Second, we could also build several condition between the connections in order to investigate how we could intervene the training process. For example, we build VAE-GAN as our UUNet architecture, then we could condition two hidden space to have certain relationship. Furthermore we gradually increase the complexity of our design with double UVAE-GAN as figure 4 (g) shows. Last, since we have double UVAE-GAN design, we tuned the proportion of discriminator's KL loss and also the ratio between the real and fake sample when forming discriminator's KL loss.
Above all, there are still wide rooms for comparison, to be simplicity we list the intra UU-Nets compared in this article as figure3. 
\subsubsection{U-like Algorithm Comparison}
We also compare with other U-Net structure methods. The original design pix2pix was compared; VAE-GAN was compared; U-like-GAN was compared and UVAE-GAN was compared.


\begin{table}
\centering
\begin{tabular}[\linewidth]{l|llll}
\hline
Model Name &  MSE   &   PSNR  &  SSIM  \\
\hline
Pix2Pix  & 0.45  & 10.64  & 0.43  \\
VAE-GAN   & 0.51  & 8.72  & 0.31 \\
Unet-VAE-GAN   & \textbf{0.40}  & \textbf{11.13}  & 0.45 \\
ours'UUNet-GAN-v1(UUnet)  & 0.52  & 8.09  & 0.44  \\
ours'UUNet-GAN-v2(UZnet)   & 0.41  & 10.74  & 0.45 \\
ours'UUNet-GAN-v3(UCnet)   & 0.49  & 7.59  & 0.43  \\
ours'UUNet-GAN-v4(fUUnet)   & 0.51  & 9.08  & 0.44  \\
ours'UUNetGAN-UUCNet  & 0.51  & 9.78 & 0.40  \\
ours'UUVAE-GAN-$Dis_{kl0.4}$   &  0.42  & 10.79
  & 0.45  \\
ours'UUVAE-GAN    &  0.44  & 10.86  & \textbf{0.47}  \\
ours'UUVAE-GAN-$Dis_{kl0.65}$ &  \textbf{0.39}  & \textbf{11.34}  & \textbf{0.49}  \\
ours'UUVAE-GAN-Tail-kl   & 0.45 & 10.59  &  0.46 \\
ours'UUVAE-GAN-ckl   & 0.43 & 10.15  &  0.50 \\
\hline
\end{tabular}
\caption{Image Translation performance on Facades dataset table, the results are averaged over 5 times computation.}
\label{tab:plain}
\end{table}

\section{Result Evaluation and Analysis}
In this section, We divide our analysis into four subsections. First, we analyze the image translation task performance. Second, we compare the results on various UU-Net designs and other state of art methods. Third, we analysis the effect by using our customized additional loss function between the discriminator and generator. Finally, we discuss stability and converged speed.
\begin{table}
\centering
\begin{tabular}{l|r|r}  
\hline

Model Name  & stability & Training Time \\
\hline

pix2pix   & 0.04  & 6h 41m      \\
UUNet-GAN-v1(UUCnet)   & \textbf{0.02}  & 11h 45m\\
UUNet-GAN-v2(UZnet)  &  0.06 &  10h 38m\\
UUNet-GAN-v3(UCnet)   & 0.04  &    16h 51m\\
UUNet-GAN-v4(FUUnet)    & 0.03  &  27h 28m\\
UUVAE-GAN-kls   & 0.05  &  13h 43m\\
\hline

\end{tabular}
\caption{Model Training Quantitative Numbers}
\label{tab:booktabs}
\end{table}
\subsection{Results on Chinese Painting and Other datasets}
We compared our methods with several state of art methods on traditional Chinese Painting novice to master painting generation and other image to image translation tasks. In order to make a fair comparison, 
we retrain all other methods from scratch with each chosen dataset, such as facades, handbag2edges and our Chinese painting dataset. The task performance quantitative results for facades dataset are showed in table 1, and the model quantitative results are showed in table 2 and figure 5. In order to measure the stability of a model, we choose the Tensorboard smoothing coefficient as one parameter in table 2. The second dimension is the duration for training 1500 epochs.

\begin{figure*}
	\begin{center}
    		\includegraphics[width=1.0\textwidth]{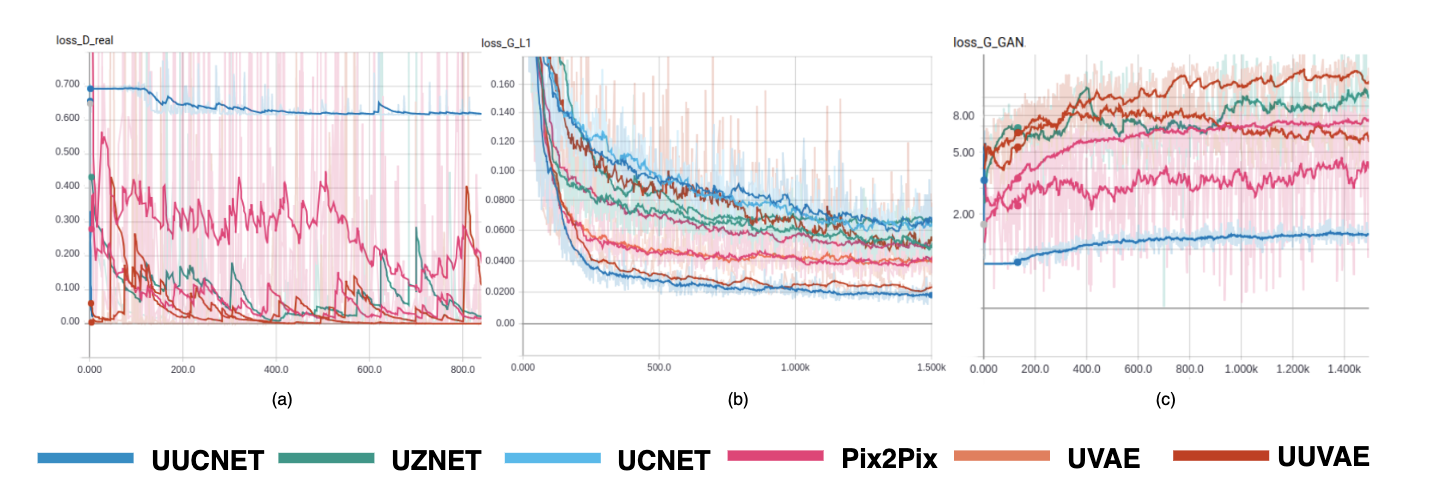}
    	\end{center}
      \caption{\small{Various networks' loss curve, the middle one is the generator's reconstruction loss, the left is the discriminator's loss with real sample and the right one is the generator's loss. For more detail loss curve can be found in supporting material. 
          	\label{fig:teaser}}}

\end{figure*}

\subsection{Result Analysis on UU-Nets Architecture}

In this section, we measure our UU-Nets architecture in terms of image translation performance. From figure 3, UU-Nets connecting the discriminator and generator, by directly carry the tensor flow from generator to discriminator, could form different connection, but the results can be quite different.
According to table 2, connecting all two components' encoder and decoder as shown in figure3 UU-Net-v4, the architecture had the best training loss curve, which force the two components' both encoders and decoders correlated. First, The connecting of encoders make sure the image located closely in terms of latent space. Second, though the function of generator's decoder and that of discriminator's is different, the gradient could flow back to generator even when we train discriminator. The reason is because we form the discriminator's decoder with generator's decoder letting the binary decision partial controlled by the generator, which not wasting the training of discriminator and makes generator surrogated in the body of discriminator. In conclusion, this mechanism mitigates the discriminator but also enhance the generator as Figure 5 (a) and (c) the blue curve shows.

To analyze the effect of the connection of different sub-nets, UUNet-GAN-v1 skip connects the encoding sub-net, which suggested that loosely combining the encoder part of discriminator and generator but not jointly making decision is showed not as good as not connecting. For UUNet-GAN-v2, it only connecting the latent code showed a little better than not connecting. For only connecting the decoding sub-net UUNet-GAN-v3, it has shown a little worse than not connecting. But in all cases, the network gradually catch up networks without connection in latter epochs. So we connect both encoders and decoders, UUNet-GAN-v4 (FUUNet) surprisingly outperform pix2pix and UVAE in early epochs during training. Besides, the tail module is crucial in stabilizing the discriminator, which adds smooth transition to the binary classification task. However our UUNet-GANs are overfitting. But we believe it is merely because of the naive UNet not matching the complexity due to the fact that UUVAEGAN group are transferred quite well.

\subsection{Analysis of UU-VAE with Restricting Loss}
As the results from the experiments, we know that the training of UU-VAE with many KL losses can be rather slow compared to other networks. In order to absorb more information, we adopted UU-UVAE with tail in figure 4 (h). Through the experiments, we noted that raising the proportion of the discriminator's KL loss could enhance the performance and vice versa showed in Table 1 and achieved the best result in test. An Interesting finding is that the best network's $\mathcal{L}_{diskl}$ on fake is weak at first but strong in latter epochs, which indicates that to train a good generator at first we make loose judgment on the fake it generated but later when it strong we also match the judgment stronger via tuning the UUVAEGAN's discriminator. Nevertheless, UU-UVAE combine all pros of our models and reached a milestone in these experiments.

\subsection{Stability and Speed Results Analysis}
In this section, we compared the duration and stability of the Networks. In Figure 5, our UUNet-GAN-v4 design reach the best result on model's stability in general are better than VAE-GAN, but it takes longer time for training. On the group of UUNet-VAE-GAN adaption we outperformed all other group of methods as shown in Table 1. The following we list the discoveries of these experiments.
\begin{enumerate}
    \item UU-UVAE's performance with higher proportion of discriminator's KL loss is accurate than that of without using it.
    \item UU-UVAE with conditional loss achieved the best performance, which not only has the UUNet family features, very small MSE, but also has the advantage in SSIM and model capacity.
    \item The result showed that UUNetGAN's training curve is good but overfitting. 
    \item UUNetGAN's discriminator is mitigated by the communication, as shown in Figure 5 that in general its discriminator loss is higher than that of pix2pix without connecting discriminator and generator.
    
\end{enumerate}

\section{Conclusion}

In this paper, we suggested a family of innovative and model-agnostic frameworks called UUNet-GANs for connecting generator and discriminator in the context of Generative Adversarial Network featuring the symmetrical design of two U-shape networks, the using of skip-connections and sharing the informative gradients between the two components during training for balancing the unfair min-max game. An intensive comparison study has been conducted for both UUNet-GAN family and other classical architectures using commonly-used image datasets, facades and Chinese painting dataset. The experiments have shown that 1) Connecting, sharing and intervening the gradient flow is an effective way to balance the min-max game; 2) a balanced adversarial training perform better; 3) UUNet-VAE model combine all the strengths of UUNets' stability and UVAE-GAN's accuracy as to image-to-image task.
\section*{Acknowledgments}
We appreciate for the precious time and efforts of the two anonymous reviewers and friends on earlier draft of this manuscript.

\section*{Acknowledgments}

The preparation of the traditional Chinese dataset and other files that implement them was supported by 
Yin Mei.
And thanks to my former colleague Shichao, who help me organize the GAN talks, and provide useful questions which is essentially help us to understand the overall architecture of generative adversarial networks.
Dr.TangXin also join the talk and have nice discussion and helped in experiment settings and solving coding issues.

{\small
\bibliographystyle{ieee}
\bibliography{paperbib}
}

\end{document}